\documentclass[10pt,twocolumn,letterpaper]{article}

\usepackage{ijcb}
\usepackage{times}
\usepackage{epsfig}
\usepackage{graphicx}
\usepackage{amsmath}
\usepackage{amssymb}
\usepackage{multirow}
\usepackage{adjustbox}
\usepackage{mathabx}
\usepackage[table]{xcolor}
\ijcbfinalcopy 

\ifijcbfinal\fi
\begin{document}

\title{Synthetic Data for the Mitigation of Demographic Biases in Face Recognition}

\author{Pietro Melzi$^1$
\and
Christian Rathgeb$^{2,3}$
\and
Ruben Tolosana$^1$
\and
Ruben Vera-Rodriguez$^1$
\and
Aythami Morales$^1$
\and
Dominik Lawatsch$^2$
\and
Florian Domin$^2$
\and
Maxim Schaubert$^2$
\and
{\tt\small $^1$Biometrics and Data Pattern Analytics Laboratory, Universidad Autonoma de Madrid, Spain}
\and
{\tt\small $^2$secunet Security Networks AG, Essen, Germany}
\and
{\tt\small $^3$Hochschule Darmstadt, Germany}
}

\maketitle
\thispagestyle{empty}

\begin{abstract}
   This study investigates the possibility of mitigating the demographic biases that affect face recognition technologies through the use of synthetic data. Demographic biases have the potential to impact individuals from specific demographic groups, and can be identified by observing disparate performance of face recognition systems across demographic groups. They primarily arise from the unequal representations of demographic groups in the training data. In recent times, synthetic data have emerged as a solution to some problems that affect face recognition systems. In particular, during the generation process it is possible to specify the desired demographic and facial attributes of images, in order to control the demographic distribution of the synthesized dataset, and fairly represent the different demographic groups. We propose to fine-tune with synthetic data existing face recognition systems that present some demographic biases. We use synthetic datasets generated with GANDiffFace, a novel framework able to synthesize datasets for face recognition with controllable demographic distribution and realistic intra-class variations. We consider multiple datasets representing different demographic groups for training and evaluation. Also, we fine-tune different face recognition systems, and evaluate their demographic fairness with different metrics. Our results support the proposed approach and the use of synthetic data to mitigate demographic biases in face recognition. 
\end{abstract}

\section{Introduction}

\begin{figure}[t]
\begin{center}
   \includegraphics[width=\linewidth]{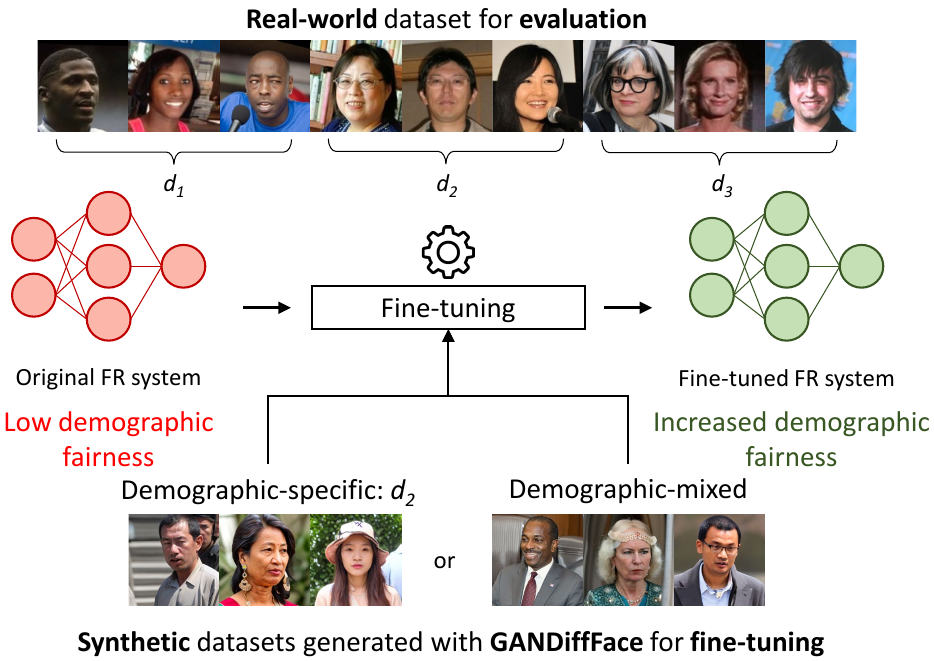}
\end{center}
   \caption{Overview of the proposed approach: Face Recognition (FR) systems with low demographic fairness are fine-tuned with demographic-specific or demographic-mixed datasets synthesized with GANDiffFace \cite{melzi2023gandiffface}. Real-world datasets representing different demographic groups (\emph{e.g.,} $d_1$ = African, $d_2$ = Asian, $d_3$ = Caucasian) are used to assess whether fairness increases in the fine-tuned FR systems.}
\label{fig:graphical}
\end{figure}

In the past few decades, face recognition technologies have spread rapidly, with applications in the scenarios of video surveillance, criminal identification, building access control, and autonomous driving to name just a few examples \cite{kortli2020face}. However, recent works have shown that numerous face recognition systems fail to provide consistent recognition performance when operating with groups of individuals of different race, gender, and age \cite{grother2022face}. As a result, face recognition systems provide different error rates for different demographic groups, causing disproportionate consequences for individuals in certain groups \cite{rathgeb2022demographic}. We consider face recognition systems to be ``unfair'' when they are biased against certain demographic groups, \emph{i.e.,} when their goodness considering the full set of data is significantly larger than their goodness considering the subset of data corresponding to the demographic groups \cite{serna2020algorithmic}. Usually, demographic biases arise unintentionally during the development of face recognition systems, with the unequal representation of demographic groups in the training data which is one of the main reasons \cite{serna2021insidebias, terhorst2021comprehensive}.

Achieving equal representation of demographic groups in the training data poses a considerable challenge. The availability of data is disproportionately higher for certain demographic groups, as evident in widely-used large-scale datasets for face recognition training \cite{Cao18, guo2016ms, yi2014learning}, and the aggregation of data from diverse sources to represent various demographic groups may not be ideal, as it introduces issues of disparate data quality across demographic groups in the final dataset \cite{xiong2018asian}. In this context, synthetic data have emerged as a promising alternative to real-world data, offering greater control over demographic and facial attributes during the synthesis process. This control enables the generation of novel synthetic datasets that align with desired demographic distributions, while maintaining consistent image quality across diverse demographic groups. Moreover, synthetic face generators have the potential to provide virtually infinite data. This is of particular importance in light of privacy concerns that have led to the discontinuation of established datasets \cite{Exposing_ai}, and regulatory frameworks like the EU-GDPR that require informed consent prior to the collection and usage of personal data \cite{voigt2017eu}.

\begin{figure}[t]
\begin{center}
   \includegraphics[width=\linewidth]{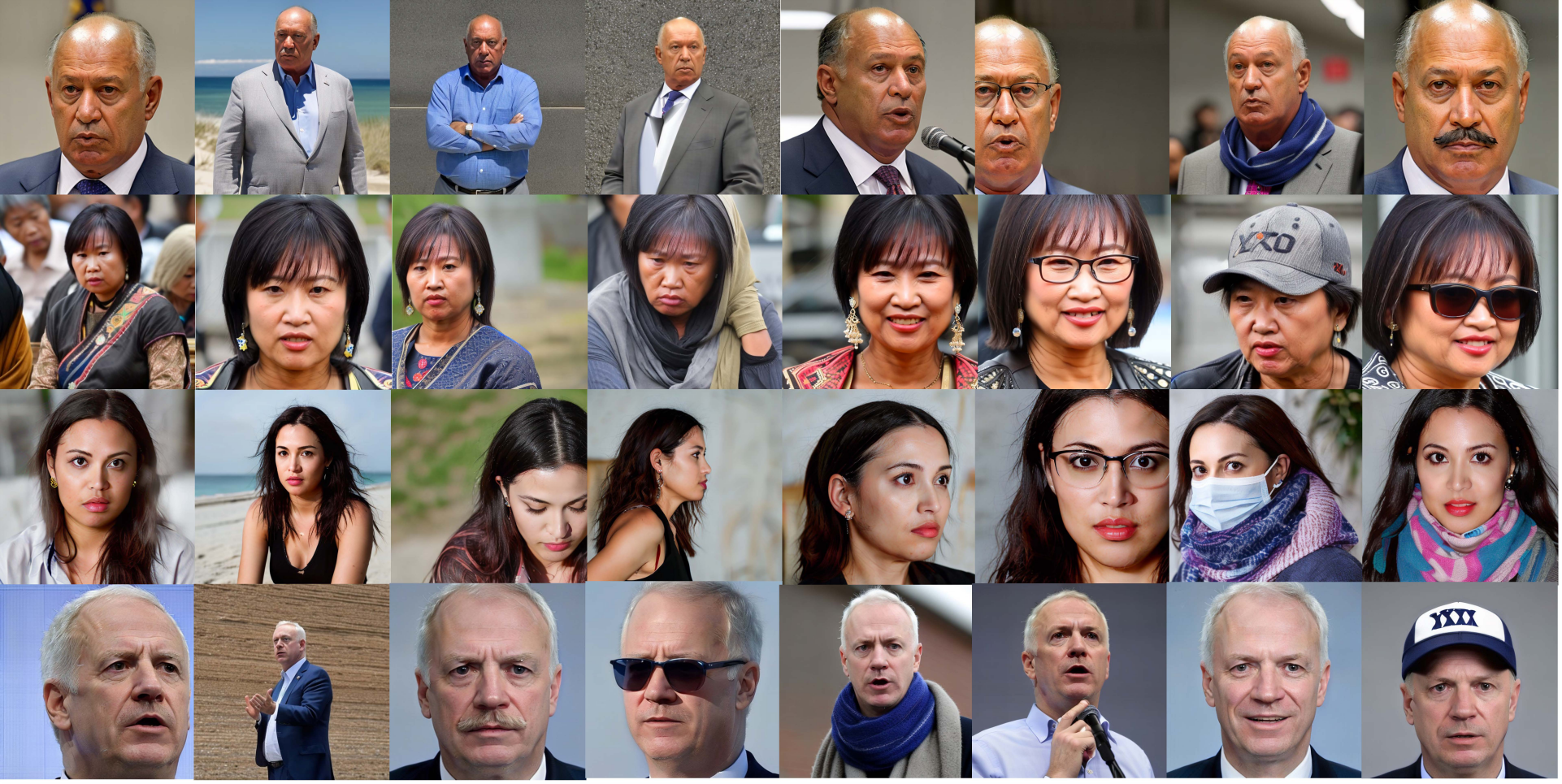}
\end{center}
   \caption{Identities (one for each row) and intra-class variations generated with GANDiffFace \cite{melzi2023gandiffface} for different demographic groups.}
\label{fig:examples}
\end{figure}

Given these inherent properties that make synthetic datasets well-suited for training face recognition systems, we investigate the application of synthetic data for the mitigation of demographic biases in face recognition. We assess the performance of widely-used face recognition systems across demographic groups, and subsequently fine-tune them using synthetic datasets designed to exhibit desired demographic distributions. The effectiveness of the proposed approach in mitigating bias is evaluated by analyzing the performance of the fine-tuned face recognition systems across demographic groups. In Figure \ref{fig:graphical} we provide an overview of the proposed approach. For the fine-tuning process, we use synthetic data generated with GANDiffFace \cite{melzi2023gandiffface}, a novel framework that synthesizes datasets possessing the properties of human face realism, controllable demographic distributions, and realistic intra-class variations. Previous studies have identified the existence of a performance gap between face recognition systems trained with synthetic and real data \cite{kortylewski2019analyzing, wang2023boosting}, attributing this disparity to the limited intra-class variation provided by synthetic data \cite{qiu2021synface}. To avoid this limitation, we consider datasets synthesized with GANDiffFace, which successfully approximate the intra-class variations observed in databases with real data \cite{melzi2023gandiffface}. Visual examples illustrating the identities and variability provided by GANDiffFace are depicted in Figure \ref{fig:examples}. To comprehensively evaluate our proposal, we employ two distinct face recognition systems, \emph{i.e.,} ArcFace \cite{deng2019arcface} and CosFace \cite{wang2018cosface}, calculate three different fairness metrics, and consider various demographic groups in both the fine-tuning and evaluation datasets. We identify the different demographic groups according to the labels provided by each dataset. This explains the concurrent use of terms like ``White'' and ``Caucasian'', or ``Black'' and ``African''.

The remainder of this work is organized as follows: Section \ref{sec:2} reviews previous studies concerning demographic bias and the utilization of synthetic data in face recognition. Section \ref{sec:3} presents the proposed method for mitigating bias and describes the fairness metrics used for evaluation. In Section \ref{sec:4}, detailed information is provided regarding the datasets and experimental configurations utilized in this study. Section \ref{sec:5} presents and discusses the results obtained. Finally, Section \ref{sec:6} draws the conclusions of this work.

\section{Related Works}
\label{sec:2}

\subsection{Demographic Biases in Face Recognition} 

Face recognition systems inherently exhibit biases that contribute to discriminatory performance variations based on demographic attributes of subjects \cite{terhorst2021comprehensive}. These biases primarily stem from training datasets, which often inadequately represent diverse demographic groups. In popular large-scale training datasets like CASIA-WebFace \cite{yi2014learning}, VGGFace2 \cite{Cao18}, and MS-Celeb-1M \cite{guo2016ms}, male, white, and middle-aged individuals are disproportionately overrepresented compared to other demographic groups. Consequently, face recognition systems trained on these data unintentionally replicate these biases, leading to substantial performance disparities among demographic groups \cite{wang2021deep}. Similar bias patterns are observed in Chinese commercial APIs, such as Face++ and Baidu, which demonstrate higher accuracy in recognizing the Asian population \cite{Wang_2019_ICCV}. 
In \cite{terhorst2021comprehensive}, various studies that investigate and identify demographic biases in face recognition are comprehensively discussed. Moreover, to quantitatively assess performance variations across demographic groups, several fairness metrics have been proposed. These metrics (described in details in Section \ref{subsec:fairness}) aim to provide a concise score indicating the fairness of face recognition systems in terms of their performance across demographic groups \cite{de2021fairness, grother2021demographic, howard2022evaluating}. In order to evaluate face recognition systems, balanced test datasets have been introduced, which represent different demographic groups with an equal number of subjects. Examples of such datasets include RFW \cite{Wang_2019_ICCV} and BFW \cite{robinson2020face}.

\subsection{Bias Mitigation}

Initial attempts at mitigating bias have primarily focused on training face recognition systems using large-scale balanced datasets. One such dataset, BUPT-BalancedFace \cite{wang2020mitigating}, comprises $1.3$ million images from $28,000$ subjects, equally representing four demographic groups: Caucasian, Indian, Asian, and African. However, it is important to note that BUPT-BalancedFace includes images either sourced from MS-Celeb-1M \cite{guo2016ms} or directly downloaded from websites, giving rise to concerns regarding subject privacy. Additionally, it is crucial to observe that even in balanced datasets, other inequalities among demographic groups may exist, such as varying numbers of images per subject, disparate image quality, poor annotations, and confounding variables entangled with group membership \cite{cherepanova2022deep}. Therefore, simply relying on balanced training datasets does not guarantee the complete resolution of bias-related issues. Certain demographic groups may require to be represented with more data, and determining the optimal representation for each demographic group is challenging.

Other measures have been proposed to mitigate bias in face recognition, including adversarial learning \cite{gong2020jointly, kim2019learning}, margin-based approaches \cite{huang2019deep}, and the analysis of a wider range of attributes beyond demographics \cite{terhorst2021comprehensive}. However, it may not be possible for algorithms to perform equally well across demographic groups. Which inequalities count as morally unacceptable bias is an ethical question, whose answer varies according to decision contexts \cite{holm2023bias}. In this study, we propose to fine-tune face recognition systems using datasets providing desired demographic representations, an approach that has received limited attention in the literature. For instance, demographic biases concerning age have been investigated in \cite{albiero2020does}, with multiple datasets of young, middle, and old people used to fine-tune face recognition systems. However, all the experiments increased the disparity of performance between different demographic groups. In \cite{vera2019facegenderid}, the gender-specific fine-tuning of face recognition systems has been successfully proposed to improve recognition performance for both males and females.

\subsection{Synthetic data in Face Recognition}
The use of synthetic datasets for the evaluation of face recognition systems has been first investigated in \cite{zhang2021applicability}, to compensate for the lack of publicly available large-scale test datasets, and in \cite{colbois2021use}, to examine the possibility of benchmarking with synthetic data. Novel technologies have recently been proposed to synthesize datasets suitable for the training of face recognition systems, based on Generative Adversarial Networks (GANs) \cite{boutros2022sface, qiu2021synface}, 3D models \cite{bae2023digiface}, and Diffusion models \cite{kim2023dcface}. Although these synthetic datasets have shown promising results, some of them present limitations that affect the performance of face recognition systems, compared to the ones trained with real-world data. 
In particular, datasets synthesized with GANs provide limited representations of intra-class variations, and datasets synthesized with 3D models lack of realism. 
Finally, synthetic datasets based on Diffusion models are promising, but still at a primitive stage. 
Recently GANDiffFace \cite{melzi2023gandiffface}, the novel framework employed in this study, has been proposed for the synthesis of datasets for face recognition. It combines the power of GANs and Diffusion models to improve face realism and demographic distributions with the former, and generate realistic intra-class variations with the latter.

\section{Proposed Methods}
\label{sec:3}
This study aims to evaluate the effectiveness of synthetic data in mitigating demographic biases in face recognition systems. The proposed approach is represented in Figure \ref{fig:graphical} and described in the following points:
\begin{enumerate}
    \item \emph{Bias identification:} using real-world evaluation datasets that contain identities representing different demographic groups, we compute the fairness metrics detailed in Section \ref{subsec:fairness}. These metrics provide an aggregate score, measuring the overall fairness of the evaluated face recognition systems. Furthermore, the analysis of the metrics calculated for each demographic group allows to identify specific demographic groups that may be affected by bias.
    \item \emph{Fine-tuning of face recognition system:} according to the results obtained at the previous step, we generate synthetic datasets to fine-tune the face recognition systems. We explore two modalities for representing demographic groups within the synthetic datasets: \emph{i)} one ensures equal representation of different demographic groups, providing a balanced approach, and \emph{ii)} the other deliberately over-represents the demographic groups that were identified as affected by bias. The objective is to determine whether these two datasets effectively mitigate bias and promote fair performance across all demographic categories.
    \item \emph{Evaluation of bias mitigation:} using the same datasets employed for bias identification in the initial step, we calculate the fairness metrics with the fine-tuned face recognition systems. By comparing the fairness scores obtained from the previous evaluation with the new scores, we can assess to which extent bias has been reduced through the process of fine-tuning the face recognition system with synthetic data.
\end{enumerate}

\subsection{Fairness Metrics}
\label{subsec:fairness}
We evaluate face recognition systems with an experimental protocol based on lists of mated and non-mated comparisons, from which false match rate (FMR) and false non-match rate (FNMR) at fixed thresholds are calculated. In particular, we fix three thresholds: $$t_1 = t_{\mathit{FMR}=10\%},\; t_2 = t_{\mathit{FMR}=1\%},\; t_3 = t_{\mathit{FMR}=0.1\%},$$ \emph{i.e.,} the thresholds that provide respectively the operational points $\mathit{FMR}=10\%$, $\mathit{FMR}=1\%$, and $\mathit{FMR}=0.1\%$, calculated on the entire set of comparisons. 

We divide mated and non-mated comparisons in the different demographic groups represented by the dataset used for evaluation. It is important to remark that the pairs of non-mated comparisons belong to the same demographic group, as non-mated comparisons of this type are the most suitable to evaluate demographic fairness. For each demographic group $d_i \in \mathbb{D}$, with $\mathbb{D} = \left\{d_1, d_2, \ldots d_n \right\}$ representing the entire set of $n$ demographic groups considered in the evaluation dataset, and each threshold $t_z \in \left\{t_1, t_2, t_3 \right\}$, we calculate all the possible $\mathit{FMR}_{d_i}@t_z$ and $\mathit{FNMR}_{d_i}@t_z$. To be considered ``fair'', a face recognition system should ideally exhibit $\mathit{FMR}_{d_i}@t_z$ around the operational point of $t_z$ for each demographic group $d_i$, and similar $\mathit{FNMR}_{d_i}@t_z$ among the different demographic groups \cite{de2021fairness}.   

Finally, to obtain an overall fairness score for the face recognition systems by aggregating the FMR and FNMR metrics across demographic groups, we calculate the three fairness metrics described in the following. These metrics aim to quantify the performance disparity between demographic groups. Each metric incorporates a parameter $\alpha \in \left(0, 1\right)$ to determine the level of concern assigned to the two different components of the formula, representing respectively a function of FMRs and FNMRs. In our study, we fix $\alpha = 0.5$ for each metric, providing equal weighting to the components.

\paragraph{Fairness Discrepancy Rate (FDR).} FDR has been proposed by scientists at the Idiap Research Institute \cite{de2021fairness}. This metric combines the maximum differences in FMRs and FNMRs between demographic groups at a given threshold $t_z$:
$$A\left(t_z\right) = \max_{d_i, d_j \in \mathbb{D}} \left( \left| \mathit{FMR}_{d_i}@t_z - \mathit{FMR}_{d_j}@t_z \right| \right)$$
$$B\left(t_z\right) = \max_{d_i, d_j \in \mathbb{D}} \left( \left| \mathit{FNMR}_{d_i}@t_z - \mathit{FNMR}_{d_j}@t_z \right| \right)$$

These differences are combined as follows:
$$\mathit{FDR}\left(t_z\right) = 1 - \left( \alpha A\left(t_z\right) + \left(1 - \alpha \right) B \left(t_z\right)\right)$$

The resulting FDR metric is on a scale of 0 to 1, with 1 being ``fair'' and 0 being ``unfair''. 

\paragraph{Inequity Rate (IR).} IR has been proposed by scientists at NIST \cite{grother2021demographic}. This metric combines the maximum ratios in FMRs and FNMRs between demographic groups at a given threshold $t_z$: 
$$A\left(t_z\right) = \frac{max_{di \in \mathbb{D}}\mathit{FMR}_{d_i}@t_z}{min_{dj \in \mathbb{D}}\mathit{FMR}_{d_j}@t_z}$$
$$B\left(t_z\right) = \frac{max_{di \in \mathbb{D}}\mathit{FNMR}_{d_i}@t_z}{min_{dj \in \mathbb{D}}\mathit{FNMR}_{d_j}@t_z}$$

These ratios are combined as follows:
$$\mathit{IR}\left(t_z\right) = A^\alpha\left(t_z\right) B^{\left( 1-\alpha \right)}\left(t_z\right)$$

The resulting IR metric is on a scale of 0 to infinite. The smaller the value is the more ``fair'' the system would be.
 
\paragraph{Gini Aggregation Rate for Biometric Equitability (GARBE).} 
GARBE is a metric proposed in \cite{howard2022evaluating} to overcome the limitations of the two other metrics proposed in the literature, \emph{i.e.,} FDR and IR. The former does not scale the values of FMRs and FNMRs to the same order of magnitude, and the latter has no theoretical upper bound and may have denominator equal to zero. 

GARBE is inspired to the mathematics of the Gini coefficient, computed for $x = \left\{\mathit{FMR}, \mathit{\mathit{FNMR}} \right\}$ as:
$$G_{x}\left(t_z\right) = \left(\frac{n}{n-1}\right) \left(\frac{\sum_{i=1}^{n} \sum_{j=1}^{n} \left| r_i - r_j \right|}{2n^2\overline{r}}\right),$$ with $r_i, r_j \in \left\{ \mathit{FMR}_{d_i}@t_z \middle| d \in \mathbb{D}  \right\}$ for $x = \mathit{FMR}$, $r_i, r_j \in \left\{ \mathit{FNMR}_{d_i}@t_z \middle| d \in \mathbb{D}  \right\}$ for $x = \mathit{FNMR}$, and $n$ the number of demographic groups.

These Gini coefficients are combined as follows: 
$$\mathit{GARBE}\left(t_z\right) = \alpha G_{\mathit{FMR}}\left(t_z\right) + \left(1 - \alpha \right)G_{\mathit{FNMR}}\left(t_z\right)$$

The resulting GARBE metric is on a scale of 0 to 1, with 0 being ``fair'' and 1 being ``unfair''.

\section{Experimental Setup}
\label{sec:4}

\subsection{Face Recognition Systems}
\label{subsec_label}
In our experimental setup, we evaluate the fine-tuning of two face recognition systems widely used in the literature: \emph{i)} ArcFace \cite{deng2019arcface} and \emph{ii)} CosFace \cite{wang2018cosface}. These Deep Neural Networks with multiple layers of feature extraction represent the state-of-the-art in face recognition. They have gained prominence due to the design of increasingly effective margin loss functions, able to generate highly discriminating features from face images \cite{wang2021deep}.
For ArcFace, we use a model pre-trained on CASIA-WebFace \cite{yi2014learning} with ResNet-18 as backbone\footnote{https://github.com/ronghuaiyang/arcface-pytorch}.
For CosFace, we use a model pre-trained on CASIA-WebFace \cite{yi2014learning} with Sphere20 as backbone\footnote{https://github.com/MuggleWang/CosFace\_pytorch}. According to \cite{wang2021deep}, these are the demographic distributions of CASIA-WebFace: 41.1\% Female, 58.9\% Male, and 84.5\% Cuacasian, 2.6\% Asian, 1.6\% Indian, 11.3\% African.

To fine-tune our face recognition systems without affecting their overall performance, we consider a conservative approach in each experiment. Specifically, we fine-tune all the layers with small learning rate of $10^{-5}$, and limit the number of training epochs to $15$.

\subsection{Evaluation Datasets}
For the final evaluation of the models, we consider two real-world datasets proposed in the literature to assess demographic fairness in face recognition: \emph{i)} DiveFace \cite{morales2020sensitivenets} and \emph{ii)} Racial Faces in-the-Wild (RFW) \cite{Wang_2019_ICCV}. DiveFace equally represents six classes obtained from the combination of gender (M, W) with three ethnic groups: Asian (A), Black (B), and Caucasian (C). It contains $24,000$ identities ($4,000$ for each class) taken from the Megaface dataset \cite{kemelmacher2016megaface}, with 5.5 images per identity in average. We randomly select one mated and one non-mated comparisons for each identity, for a total of $48,000$ image pairs for evaluation. 
RFW contains four testing subsets, namely Caucasian (CA), Asian (AS), Indian (IN) and African (AF). 
Each testing subset contains about $3,000$ individuals and provides an evaluation protocol with $6,000$ image pairs for face verification, for a total of $24,000$ image pairs.

\begin{figure}[t]
\begin{center}
   \includegraphics[width=\linewidth]{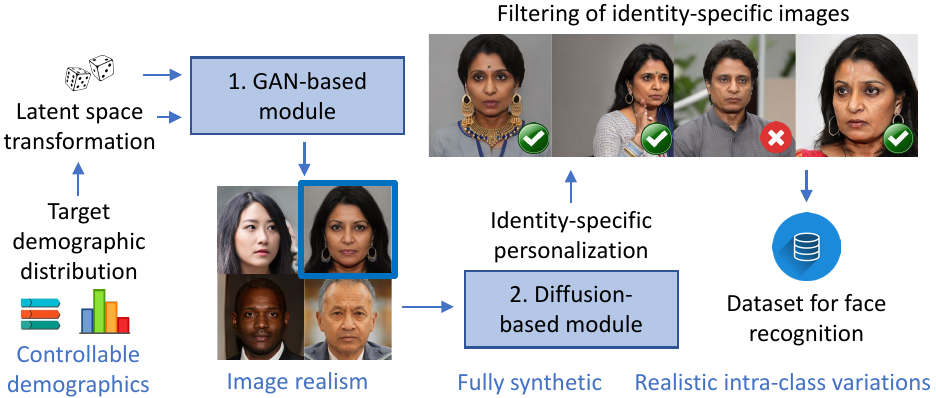}
\end{center}
   \caption{Overview of GANDiffFace \cite{melzi2023gandiffface} based on the combination of GAN and Diffusion models. GANDiffFace creates synthetic datasets for face recognition with the properties listed in blue.}
\label{fig:gandiffface}
\end{figure}

\subsection{Fine-Tuning Datasets}
For fine-tuning, we consider synthetic face datasets generated with GANDiffFace \cite{melzi2023gandiffface}, a recently proposed framework that combines the power of GANs and Diffusion models to synthesize datasets for face recognition. An overview of the GANDiffFace framework is provided in Figure \ref{fig:gandiffface}. With GANDiffFace, from each identity synthesized with the GAN-based module, a personalized Diffusion-based module generates images with realistic intra-class variations that, once filtered, will compose the final dataset. GANDiffFace allows to generate synthetic identities that represent subjects with specific ethnicity, gender, and age interval. The synthetic images included in the dataset present variations of pose, expression, illumination, and contain several occlusions, \emph{e.g.,} sunglasses, spectacles, hands, masks, scarfs, microphones, make-up, hats.

When evaluating the face recognition systems using both the evaluation datasets, \emph{i.e.,} DiveFace and RFW, we observe a bias against the Asian population in both ArcFace and CosFace (see Section \ref{sec:5}). Hence, we synthesize two datasets with GANDiffFace to fine-tune our face recognition systems: \emph{i)} the Syn-Asian dataset and \emph{ii)} the Syn-Mixed dataset. The Syn-Asian dataset contains $200$ synthetic identities ($44$ images per identity) representing Asian subjects with balanced gender and age. The Syn-Mixed dataset contains $300$ synthetic identities ($38$ images per identity) representing Asian, Black, and White subjects with balanced gender and age. With MagFace \cite{meng2021magface} we assess the image quality across demographic groups in the Syn-Mixed dataset, observing similar qualities for the three demographic groups represented in the dataset: $27.37$ ($\pm1.78$) for Asian, $26.98$ ($\pm1.62$) for Black, and $26.20$ ($\pm1.50$) for White populations.

\section{Results}
\label{sec:5}
In our experiments, we perform fine-tuning on two face recognition systems, \emph{i.e.,} ArcFace and CosFace, employing two distinct datasets synthesized with GANDiffFace, \emph{i.e.,} Syn-Asian and Syn-Mixed, and evaluating the performance of both the original and fine-tuned systems using two real-world datasets, \emph{i.e.,} DiveFace and RFW. The results obtained for the ArcFace model are reported in Tables \ref{tab:1}, \ref{tab:2}, and \ref{tab:3}, and further discussed in Section \ref{subsec:arcface}. Similarly, the results obtained for the CosFace model are reported in Tables \ref{tab:4}, \ref{tab:5}, and \ref{tab:6}, and further discussed in Section \ref{subsec:cosface}. These tables provide comprehensive insights into the performance metrics obtained from each experiment, highlighting the impact of fine-tuning with different synthetic datasets.

We now present some general observations that apply to all experiments conducted. It is evident that subjects from the DiveFace dataset are relatively easier to recognize compared to those from the RFW dataset, as demonstrated by the FNMRs reported in Tables \ref{tab:1} and \ref{tab:4} for DiveFace, and Tables \ref{tab:2} and \ref{tab:5} for RFW. Specifically, at the middle threshold $t_2$, we achieve FNMRs ranging between $3$ and $5\%$ for DiveFace, while for RFW, the FNMRs range between $61$ and $88\%$. It is worth noting that RFW considers both the large intra-class variance and the tiny inter-class variance to avoid saturated performance \cite{Wang_2019_ICCV}. 
Furthermore, in both ArcFace and CosFace models, we observe higher FMRs for the Asian population (AM and AW groups in DiveFace, AS group in RFW) compared to the other demographic groups. We especially focus on FMR as opposed to FNMR, as the latter can be strongly influenced by image quality, with poor quality images that can induce a demographic effect \cite{grother2022face}. Higher FMRs align with our expectations, as the ArcFace and CosFace models utilized in our experiments were trained on the CASIA-WebFace dataset \cite{yi2014learning}, where only $2.6\%$ of the subjects are Asian.

\subsection{ArcFace}

\begin{table}[]
\begin{center}
\begin{tabular}{|c|ccc|ccc|}
\hline
\multirow{3}{*}{\textbf{DG}} & \multicolumn{3}{c|}{FMR $\left[\%\right]$}                                                             & \multicolumn{3}{c|}{FNMR $\left[\%\right]$}                               \\ \cline{2-7} 
                                            & \multicolumn{1}{c|}{$t_1$} & \multicolumn{1}{c|}{$t_2$} & \multicolumn{1}{c|}{$t_3$} & \multicolumn{1}{c|}{$t_1$} & \multicolumn{1}{c|}{$t_2$} & \multicolumn{1}{c|}{$t_3$} \\ \cline{2-7}
                                            & \multicolumn{6}{c|}{\textbf{ArcFace original}}                                                                                             \\ \hline 
\textbf{AM} & \textbf{20.03} & \textbf{2.48} & \textbf{0.25} & 0.53 & 2.10 & 5.55 \\
\textbf{AW} & \textbf{18.27} & \textbf{2.43} & \textbf{0.25} & 0.40 & 1.85 & 6.12 \\
BM & 5.67 & 0.20 & 0.00 & 0.68 & 3.28 & 8.82 \\
BW & 7.25 & 0.57 & 0.07 & 1.47 & 4.52 & 9.15 \\
CM & 3.28 & 0.07 & 0.00 & 0.80 & 3.20 & 9.10 \\
CW & 5.50 & 0.25 & 0.03 & 0.60 & 2.90 & 9.57 \\ \hline
OV & 10.00 & 1.00 & 0.10 & 0.75 & 2.98 & 8.05 \\ \hline

& \multicolumn{6}{|c|}{\textbf{ArcFace Syn-Asian FT}}                                                                                             \\ \hline 
\textbf{AM} & \cellcolor{red!25} 1.43 & \cellcolor{red!25} 0.05 & \cellcolor{red!25} 0.00 & 3.55 & 14.00 & 28.78 \\
\textbf{AW} & \cellcolor{red!25} 1.52 & \cellcolor{red!25} 0.07 & \cellcolor{red!25} 0.00 & 2.90 & 13.30 & 30.08 \\
BM & 14.10 & 1.38 & 0.22 & 0.60 & 4.62 & 13.70 \\
BW & 6.93 & 0.40 & 0.00 & 2.10 & 7.83 & 17.72 \\
CM & 18.12 & 2.00 & 0.18 & 0.55 & 2.65 & 7.92 \\
CW & 17.90 & 2.10 & 0.20 & 0.38 & 3.17 & 10.03 \\ \hline
OV & 10.00 & 1.00 & 0.10 & 1.68 & 7.60 & 18.04 \\ \hline

& \multicolumn{6}{|c|}{\textbf{ArcFace Syn-Mixed FT}}                                                                                             \\ \hline 
                                            
\textbf{AM} & \cellcolor{green!25} 11.35 & \cellcolor{green!25} 1.18 & \cellcolor{green!25} 0.05 & 0.90 & 3.92 & 9.72 \\
\textbf{AW} & \cellcolor{green!25} 12.60 & \cellcolor{green!25} 1.30 & \cellcolor{green!25} 0.15 & 0.60 & 3.33 & 8.22 \\
BM & 8.18 & 0.75 & 0.10 & 0.62 & 3.80 & 10.45 \\
BW & 8.95 & 0.88 & 0.03 & 1.65 & 5.05 & 11.18 \\
CM & 8.43 & 0.75 & 0.15 & 0.70 & 2.53 & 6.88 \\
CW & 10.50 & 1.15 & 0.12 & 0.55 & 3.08 & 8.60 \\ \hline
OV & 10.00 & 1.00 & 0.10 & 0.84 & 3.62 & 9.18 \\ \hline

\end{tabular}
\end{center}
\caption{Evaluation of ArcFace with DiveFace dataset. AM = Asian Man, AW = Asian Woman, BM = Black Man, BW = Black Woman, CM = Caucasian Man, CW = Caucasian Woman, DG = Demographic Group, FT = Fine-Tuned, OV = Overall.}
\label{tab:1}
\end{table}

\begin{table}[]
\begin{center}
\begin{tabular}{|c|ccc|ccc|}
\hline
\multirow{3}{*}{\textbf{DG}} &  \multicolumn{3}{c|}{FMR $\left[\%\right]$}                                                             & \multicolumn{3}{c|}{FNMR $\left[\%\right]$}                               \\ \cline{2-7} 
                                            & \multicolumn{1}{c|}{$t_1$} & \multicolumn{1}{c|}{$t_2$} & \multicolumn{1}{c|}{$t_3$} & \multicolumn{1}{c|}{$t_1$} & \multicolumn{1}{c|}{$t_2$} & \multicolumn{1}{c|}{$t_3$} \\ \cline{2-7}
                                            & \multicolumn{6}{c|}{\textbf{ArcFace original}}                                                                                             \\ \hline
AF & 13.33 & 1.10 & 0.13 & 30.93 & 66.27 & 83.97 \\
\textbf{AS} & \textbf{15.93} & \textbf{1.83} & \textbf{0.17} & 27.47 & 57.30 & 72.97 \\
CA & 1.30 & 0.00 & 0.00 & 36.40 & 66.23 & 81.60 \\
IN & 9.43 & 1.07 & 0.10 & 27.27 & 56.00 & 74.10 \\ \hline
OV & 10.00 & 1.00 & 0.10 & 30.52 & 61.45 & 78.16 \\ \hline

& \multicolumn{6}{|c|}{\textbf{ArcFace Syn-Asian FT}}                                                                                             \\ \hline 
AF & 27.30 & 3.23 & 0.37 & 33.77 & 73.63 & 91.47 \\
\textbf{AS} & \cellcolor{red!25} 0.53 & \cellcolor{red!25} 0.00 & \cellcolor{red!25} 0.00 & 73.57 & 92.87 & 97.67 \\
CA & 5.67 & 0.43 & 0.00 & 34.87 & 71.33 & 88.67 \\
IN & 6.50 & 0.33 & 0.03 & 42.23 & 76.60 & 91.97 \\ \hline
OV & 10.00 & 1.00 & 0.10 & 46.11 & 78.61 & 92.44 \\ \hline

& \multicolumn{6}{|c|}{\textbf{ArcFace Syn-Mixed FT}}                                                                                             \\ \hline 
                                            
AF & 14.07 & 1.43 & 0.13 & 40.73 & 75.63 & 90.07 \\
\textbf{AS} & \cellcolor{green!25} 9.23 & \cellcolor{green!25} 0.73 & \cellcolor{green!25} 0.07 & 41.00 & 72.47 & 86.00 \\
CA & 4.13 & 0.23 & 0.03 & 34.53 & 65.97 & 82.50 \\
IN & 12.57 & 1.60 & 0.17 & 30.73 & 61.83 & 79.83 \\ \hline
OV & 10.00 & 1.00 & 0.10 & 36.75 & 68.98 & 84.60 \\ \hline

\end{tabular}
\end{center}
\caption{Evaluation of ArcFace with RFW dataset. AF = African, AS = Asian, CA = Caucasian, IN = Indian, DG = Demographic Group, FT = Fine-Tuned, OV = Overall.}
\label{tab:2}
\end{table}

\begin{table}[]
\begin{center}
\begin{tabular}{|c|ccc|ccc|}
\hline
\multirow{3}{*}{\textbf{FM}} &  \multicolumn{3}{c|}{DiveFace}                                                             & \multicolumn{3}{c|}{RFW}                               \\ \cline{2-7} 
                                            & \multicolumn{1}{c|}{$t_1$} & \multicolumn{1}{c|}{$t_2$} & \multicolumn{1}{c|}{$t_3$} & \multicolumn{1}{c|}{$t_1$} & \multicolumn{1}{c|}{$t_2$} & \multicolumn{1}{c|}{$t_3$} \\ \cline{2-7}
                                            & \multicolumn{6}{c|}{\textbf{ArcFace original}}                                                                                             \\ \hline
FDR $\uparrow$ & 0.91 & 0.97 & 0.98 & 0.88 & 0.94 & 0.94 \\
IR $\downarrow$ & 4.75 & 8.98 & inf & 4.04 & inf & inf \\
\textbf{GARBE $\downarrow$} & \textbf{0.35} & \textbf{0.41} & \textbf{0.40} & \textbf{0.24} & \textbf{0.26} & \textbf{0.24} \\ \hline

& \multicolumn{6}{|c|}{\textbf{ArcFace Syn-Asian FT}}                                                                                             \\ \hline 
FDR $\uparrow$ & 0.90 & 0.93 & 0.89 & 0.67 & 0.88 & 0.95\\
IR $\downarrow$ & 10.97 & 14.90 & inf & 10.56 & inf & inf \\
\textbf{GARBE $\downarrow$} & \cellcolor{red!25} 0.48 & \cellcolor{red!25} 0.48 & \cellcolor{red!25} 0.47 & \cellcolor{red!25} 0.45 & \cellcolor{red!25} 0.44 & \cellcolor{red!25} 0.48\\ \hline

& \multicolumn{6}{|c|}{\textbf{ArcFace Syn-Mixed FT}}                                                                                             \\ \hline 
                                            
FDR $\uparrow$ & 0.97 & 0.98 & 0.98 & 0.90 & 0.92 & 0.95 \\
IR $\downarrow$ & 2.15 & 1.86 & 3.12 & 2.13 & 2.90 & 2.38 \\
\textbf{GARBE $\downarrow$} & \cellcolor{green!25} 0.18 & \cellcolor{green!25} 0.14 & \cellcolor{green!25} 0.21 & \cellcolor{green!25} 0.18 & \cellcolor{green!25} 0.23 & \cellcolor{green!25} 0.21 \\ \hline

\end{tabular}
\end{center}
\caption{Fairness Metrics (FM) for the original and Fine-Tuned (FT) ArcFace models. Arrows indicate the direction in which metrics represent better fairness. inf = infinite.}
\label{tab:3}
\end{table}

\label{subsec:arcface}
The fine-tuning process of the original ArcFace system using the Syn-Asian dataset yields undesired outcomes during the evaluation of the resulting system. In order for a face recognition system to be considered fair, it should ideally exhibit similar FMRs across demographic groups at all the fixed operational points (\emph{i.e.,} overall $\mathit{FMR} = 10\%$, $1\%$, and $0.1\%$). However, we observe in Tables \ref{tab:1} and \ref{tab:2} that the original ArcFace system provides FMRs for the Asian population that are approximately twice as high (\emph{i.e.,} around $20\%$, $2\%$, and $0.2\%$). When the ArcFace system is fine-tuned with the Syn-Asian dataset, the resulting FMRs exhibit values approximately one-tenth of the values at operational points (\emph{i.e.,} around $1\%$, $0.1\%$, and $0.01\%$). Consequently, this discrepancy leads to a deterioration in the fairness metrics, as illustrated in Table \ref{tab:3}, with the GARBE values increasing by at least $20\%$ across different thresholds and test datasets.

In contrast, the fine-tuning of ArcFace using the Syn-Mixed dataset provides more favorable results. The fine-tuned system achieves FMRs for the Asian population that are comparable to the fixed values of $10\%$, $1\%$, and $0.1\%$, aligning with the performance of other demographic groups (as observed in Tables \ref{tab:1} and \ref{tab:2}). Additionally, we note that the fine-tuning with Syn-Mixed dataset only marginally increases the FNMRs obtained at different thresholds, differently from the fine-tuning with the Syn-Asian dataset (Tables \ref{tab:1} and \ref{tab:2}). This indicates that the overall performance of ArcFace fine-tuned with the Syn-Mixed dataset remains sufficiently similar to the original one. Finally, we observe a general improvement of the fairness metrics, with GARBE evaluated on DiveFace showing a halving of its value across different thresholds, as reported in Table \ref{tab:3}.

\subsection{CosFace}
\label{subsec:cosface}

\begin{table}[]
\begin{center}
\begin{tabular}{|c|ccc|ccc|}
\hline
\multirow{3}{*}{\textbf{DG}} &  \multicolumn{3}{c|}{FMR $\left[\%\right]$}                                                             & \multicolumn{3}{c|}{FNMR $\left[\%\right]$}                               \\ \cline{2-7} 
                                            & \multicolumn{1}{c|}{$t_1$} & \multicolumn{1}{c|}{$t_2$} & \multicolumn{1}{c|}{$t_3$} & \multicolumn{1}{c|}{$t_1$} & \multicolumn{1}{c|}{$t_2$} & \multicolumn{1}{c|}{$t_3$} \\ \cline{2-7}
                                            & \multicolumn{6}{c|}{\textbf{CosFace original}}                                                                                             \\ \hline 
\textbf{AM} & \textbf{16.90} & \textbf{1.50} & \textbf{0.10} & 0.88 & 3.40 & 8.33 \\
\textbf{AW} & \textbf{25.72} & \textbf{3.55} & \textbf{0.40} & 0.57 & 3.48 & 9.30 \\
BM & 3.20 & 0.10 & 0.00 & 1.05 & 5.88 & 14.40 \\
BW & 7.62 & 0.57 & 0.07 & 1.60 & 6.50 & 14.10 \\
CM & 1.65 & 0.03 & 0.00 & 1.23 & 6.60 & 15.80 \\
CW & 4.90 & 0.25 & 0.03 & 0.90 & 6.15 & 15.38 \\ \hline
OV & 10.00 & 1.00 & 0.10 & 1.04 & 5.33 & 12.88 \\ \hline

& \multicolumn{6}{|c|}{\textbf{CosFace Syn-Asian FT}}                                                                                             \\ \hline 
\textbf{AM} & \cellcolor{green!25} 16.45 & \cellcolor{green!25} 1.62 & \cellcolor{green!25} 0.12 & 0.80 & 3.38 & 9.00 \\
\textbf{AW} & \cellcolor{green!25} 22.88 & \cellcolor{green!25} 3.02 & \cellcolor{green!25} 0.35 & 0.70 & 3.52 & 9.68 \\
BM & 4.08 & 0.15 & 0.00 & 1.05 & 5.10 & 13.78 \\
BW & 8.70 & 0.75 & 0.07 & 1.38 & 6.05 & 13.33 \\
CM & 2.25 & 0.03 & 0.00 & 0.90 & 5.90 & 14.20 \\
CW & 5.65 & 0.43 & 0.05 & 0.80 & 5.25 & 14.30 \\ \hline
OV & 10.00 & 1.00 & 0.10 & 0.94 & 4.87 & 12.38 \\ \hline

& \multicolumn{6}{|c|}{\textbf{CosFace Syn-Mixed FT}}                                                                                             \\ \hline 
                                            
\textbf{AM} & \cellcolor{red!25} 16.65 & \cellcolor{red!25} 1.43 & \cellcolor{red!25} 0.12 & 0.70 & 3.90 & 9.68 \\
\textbf{AW} & \cellcolor{red!25} 27.22 & \cellcolor{red!25} 3.98 & \cellcolor{red!25} 0.43 & 0.53 & 3.60 & 9.57 \\
BM & 3.02 & 0.07 & 0.00 & 1.32 & 6.38 & 16.30 \\
BW & 6.58 & 0.30 & 0.03 & 1.70 & 7.70 & 16.20 \\
CM & 1.88 & 0.05 & 0.00 & 1.05 & 7.20 & 17.15 \\
CW & 4.65 & 0.18 & 0.03 & 1.03 & 7.03 & 17.42 \\ \hline
OV & 10.00 & 1.00 & 0.10 & 1.05 & 5.97 & 14.39 \\ \hline

\end{tabular}
\end{center}
\caption{Evaluation of CosFace with DiveFace dataset. AM = Asian Man, AW = Asian Woman, BM = Black Man, BW = Black Woman, CM = Caucasian Man, CW = Caucasian Woman, DG = Demographic Group, FT = Fine-Tuned, OV = Overall.}
\label{tab:4}
\end{table}

\begin{table}[]
\begin{center}
\begin{tabular}{|c|ccc|ccc|}
\hline
\multirow{3}{*}{\textbf{DG}}  & \multicolumn{3}{c|}{FMR $\left[\%\right]$}                                                             & \multicolumn{3}{c|}{FNMR $\left[\%\right]$}                               \\ \cline{2-7} 
                                            & \multicolumn{1}{c|}{$t_1$} & \multicolumn{1}{c|}{$t_2$} & \multicolumn{1}{c|}{$t_3$} & \multicolumn{1}{c|}{$t_1$} & \multicolumn{1}{c|}{$t_2$} & \multicolumn{1}{c|}{$t_3$} \\ \cline{2-7}
                                            & \multicolumn{6}{c|}{\textbf{CosFace original}}                                                                                             \\ \hline
AF & 10.30 & 0.53 & 0.03 & 68.27 & 93.80 & 98.47 \\
\textbf{AS} & \textbf{16.17} & \textbf{1.97} & \textbf{0.13} & 49.93 & 82.50 & 92.53 \\
CA & 3.53 & 0.13 & 0.00 & 61.90 & 89.77 & 96.73 \\
IN & 10.00 & 1.37 & 0.23 & 55.00 & 86.13 & 94.73 \\ \hline
OV & 10.00 & 1.00 & 0.10 & 58.78 & 88.05 & 95.62 \\ \hline

& \multicolumn{6}{|c|}{\textbf{CosFace Syn-Asian FT}}                                                                                             \\ \hline 
AF & 13.23 & 0.80 & 0.07 & 64.63 & 92.37 & 98.70 \\
\textbf{AS} & \cellcolor{green!25} 10.10 & \cellcolor{green!25} 1.23 & \cellcolor{green!25} 0.13 & 58.60 & 87.07 & 95.53 \\
CA & 5.37 & 0.17 & 0.00 & 56.17 & 86.50 & 96.93 \\
IN & 11.30 & 1.80 & 0.20 & 51.67 & 84.03 & 95.47 \\ \hline
OV & 10.00 & 1.00 & 0.10 & 57.77 & 87.49 & 96.66 \\ \hline

& \multicolumn{6}{|c|}{\textbf{CosFace Syn-Mixed FT}}                                                                                             \\ \hline 
                                            
AF & 8.27 & 0.27 & 0.00 & 72.90 & 96.07 & 98.87 \\
\textbf{AS} & \cellcolor{red!25} 15.03 & \cellcolor{red!25} 2.33 & \cellcolor{red!25} 0.27 & 53.80 & 85.63 & 93.80 \\
CA & 3.73 & 0.07 & 0.00 & 63.87 & 91.67 & 97.47 \\
IN & 12.97 & 1.33 & 0.13 & 52.83 & 86.63 & 95.17 \\ \hline
OV & 10.00 & 1.00 & 0.10 & 60.85 & 90.00 & 96.33 \\ \hline

\end{tabular}
\end{center}
\caption{Evaluation of CosFace with RFW dataset. AF = African, AS = Asian, CA = Caucasian, IN = Indian, DG = Demographic Group, FT = Fine-Tuned, OV = Overall.}
\label{tab:5}
\end{table}

\begin{table}[]
\begin{center}
\begin{tabular}{|c|ccc|ccc|}
\hline
\multirow{3}{*}{\textbf{FM}} &  \multicolumn{3}{c|}{DiveFace}                                                             & \multicolumn{3}{c|}{RFW}                               \\ \cline{2-7} 
                                            & \multicolumn{1}{c|}{$t_1$} & \multicolumn{1}{c|}{$t_2$} & \multicolumn{1}{c|}{$t_3$} & \multicolumn{1}{c|}{$t_1$} & \multicolumn{1}{c|}{$t_2$} & \multicolumn{1}{c|}{$t_3$} \\ \cline{2-7}
                                            & \multicolumn{6}{c|}{\textbf{CosFace original}}                                                                                             \\ \hline
FDR $\uparrow$ & 0.87 & 0.97 & 0.96 & 0.85 & 0.93 & 0.97 \\
IR $\downarrow$ & 6.59 & 16.60 & inf & 2.50 & 4.10 & inf \\
\textbf{GARBE $\downarrow$} & \textbf{0.38} & \textbf{0.45} & \textbf{0.46} & \textbf{0.20} & \textbf{0.28} & \textbf{0.34} \\ \hline

& \multicolumn{6}{|c|}{\textbf{CosFace Syn-Asian FT}}                                                                                             \\ \hline 
FDR $\uparrow$ & 0.89 & 0.97 & 0.97 & 0.90 & 0.95 & 0.98 \\
IR $\downarrow$ & 4.47 & 14.73 & inf & 1.76 & 3.45 & inf \\
\textbf{GARBE $\downarrow$} & \cellcolor{green!25} 0.31 & \cellcolor{green!25} 0.40 & \cellcolor{green!25} 0.41 & \cellcolor{green!25} 0.13 & \cellcolor{green!25} 0.23 & \cellcolor{green!25} 0.28 \\ \hline

& \multicolumn{6}{|c|}{\textbf{CosFace Syn-Mixed FT}}                                                                                             \\ \hline 
                                            
FDR $\uparrow$ & 0.87 & 0.96 & 0.96 & 0.84 & 0.94 & 0.97 \\
IR $\downarrow$ & 6.86 & 13.04 & inf & 1.76 & 3.45 & inf \\
\textbf{GARBE $\downarrow$} & \cellcolor{red!25} 0.41 & \cellcolor{red!25} 0.48 & \cellcolor{red!25} 0.49 & \cellcolor{red!25} 0.21 & \cellcolor{red!25} 0.34 & \cellcolor{red!25} 0.40 \\ \hline

\end{tabular}
\end{center}
\caption{Fairness Metrics (FM) for the original and Fine-Tuned (FT) CosFace models. Arrows indicate the direction in which metrics represent better fairness. inf = infinite.}
\label{tab:6}\vspace{-0.4cm}
\end{table}

In the case of the CosFace system, fine-tuning with the Syn-Asian dataset yields slight improvements in the evaluation of the resulting system, specifically for the FMRs associated with the Asian population (as shown in Tables \ref{tab:4} and \ref{tab:5}). Although the improvement may be less significant compared to the fine-tuning of ArcFace with the Syn-Mixed dataset, there is a noticeable enhancement in the fairness metrics overall. Table \ref{tab:6} demonstrates a general improvement, with the GARBE values decreasing by at least $10\%$ across different thresholds and test datasets. Notably, in contrast to the fine-tuning of ArcFace, we also observe a decrease in the overall FNMRs at different thresholds for CosFace. This indicates an overall improvement in the performance of CosFace due to the fine-tuning process.

In contrast, the fine-tuning of CosFace with the Syn-Mixed dataset does not yield any discernible benefits for the Asian populations under consideration in our evaluations. We observe an increase in the FMRs for the AW group in DiveFace and the AS group in RFW, as depicted in Tables \ref{tab:4} and \ref{tab:5}. As a consequence, the fairness metrics evaluated on the CosFace system fine-tuned with the Syn-Mixed dataset are inferior to those evaluated on the original CosFace system (Table \ref{tab:6}). This suggests that the fine-tuning process with the Syn-Mixed dataset may not be effective in improving the fairness and performance of CosFace for the Asian populations.

\section{Conclusion}
\label{sec:6}
In this study, we investigated the use of synthetic data for fine-tuning face recognition systems in order to mitigate performance discrepancies across demographic groups. Both face recognition systems examined in our experiments presented biases affecting the recognition performance for the Asian population. To address this issue, we fine-tuned these systems using two distinct synthetic datasets that provided realistic intra-class variations. The first dataset, Syn-Asian, exclusively comprised Asian subjects, while the second dataset, Syn-Mixed, encompassed subjects from three different demographic groups, \emph{i.e.,} Asian, Black, and White. Our observations revealed distinct behaviors for the two face recognition systems. For ArcFace, fine-tuning with the Syn-Asian dataset produced a significant effect, substantially reducing the FMRs of the Asian population to levels below those of other demographic groups. On the other hand, fine-tuning with the Syn-Mixed dataset improved the fairness metrics. In the case of CosFace, modifying the performance of the original system proved to be more challenging. While fine-tuning with the Syn-Mixed dataset did not adequately address the observed bias, fine-tuning with the Syn-Asian dataset resulted in improved fairness metrics and overall accuracy of the face recognition system.

Synthetic data offer numerous advantages compared to real data, and we expect a growing adoption of synthetic data in face recognition in the near future. With this study, we have demonstrated that synthetic data can be successfully employed to mitigate bias in face recognition, as they can easily represent different demographic groups exhibiting similar image quality across demographic groups and realistic intra-class variations. In the future, we also expect that synthetic data will be used for the evaluation of face recognition systems. It is plausible that, as real-world datasets potentially become discontinued, synthetic datasets may become the primary source for benchmarking face recognition technology.

\section*{Acknowledgments}
This project has received funding from the European Union’s Horizon 2020 research and innovation programme under the Marie Sk\l{}odowska-Curie grant agreement No 860813 - TReSPAsS-ETN, and it is supported by INTER-ACTION (PID2021- 126521OB-I00 MICINN/FEDER) and C\'atedra ENIA UAM-VERIDAS en IA Responsable (NextGenerationEU PRTR TSI-100927-2023-2). This research is based upon work supported by the Hessian Ministry of the Interior and Sport in the course of the Bio4ensics project and by the German Federal Ministry of Education and Research and the Hessian Ministry of Higher Education, Research, Science and the Arts within their joint support of the National Research Center for Applied Cybersecurity ATHENE.

{\small
\bibliographystyle{ieee}
\bibliography{egbib}
}

\end{document}